\documentclass{article}
\usepackage{url}
\usepackage{graphicx} % Required for inserting images
\usepackage{booktabs} 
\usepackage{authblk}
\usepackage{float}
\title{Investigating the Efficacy of Large Language Models in Reflective Assessment Methods through Chain of Thoughts Prompting}

\author{Baphumelele Masikisiki\textsuperscript{1}, Vukosi Marivate\textsuperscript{2}, Yvette Hlophe\textsuperscript{3}  \\
\textsuperscript{1} University of Pretoria,  bmasikisiki@gmail.com \\
\textsuperscript{2}University of Pretoria,  vukosi.marivate@cs.up.za \\
\textsuperscript{3}University of Pretoria,  yvette.hlophe@up.ac.za}

\date{}
\begin{document}

\maketitle

\begin{abstract}
Large Language Models, such as Generative Pre-trained Transformer 3 (aka. GPT-3), have been developed to understand language through the analysis of extensive text data, allowing them to identify patterns and connections between words. While LLMs have demonstrated impressive performance across various text-related tasks, they encounter challenges in tasks associated with reasoning. To address this challenge, \textit{Chain of Thought} (CoT) prompting method has been proposed as a means to enhance LLMs' proficiency in complex reasoning tasks like solving math word problems and answering questions based on logical argumentative reasoning. The primary aim of this research is to assess how well four language models can grade reflective essays of third-year medical students. The assessment will specifically target the evaluation of critical thinking skills using \textit{ CoT} prompting.

The research will provide the following contributions; to introduce and educate on the process of instructing models to evaluate reflective essays from a dataset they have not been previously trained on; to illustrate the use of \textit{CoT} prompting as an instructional approach for training large models to carry out particular tasks. Our results suggest that among all the models, Llama-7b performs the least effectively, displaying the highest mean squared error. Conversely, {ChatGPT} emerges as the superior model, boasting a higher Cohen kappa score value of 0.53. Lastly, it's important to note that the selected models do prioritise user privacy by allowing users to delete their own conducted conversations.
\end{abstract}

\section{Introduction}

Reflective writing \cite{gibson2017reflective} is an assessment technique that helps students demonstrate a deeper understanding of a subject. This form of writing encourages critical thinking, thereby supporting scholars in developing skills such as evaluation, synthesis, and self-regulation that foster independent learning \cite{azer2008use}. However, this assessment method relies on manual grading processes, which are considered time-intensive, costly, prone to inconsistency, and lacking scalability\cite{ullmann2017reflective}. Scholars in the field of Natural Language Processing (NLP) have been investigating automated grading methods for this writing style for quite some time \cite{beigman2017reflective}. However, automated grading for long texts such as essays, and reflective writing remains partially resolved because of challenges, such as the inability to effectively handle lengthy texts and provide students with a well-justified score \cite{ramesh2022automated}.  Well-justified score \cite{ramesh2022automated} talks about the ability to provide both students and teachers with informative feedback for the assigned score.

However, the emergence of Large Language Models (LLMs) \cite{strasser2023pitfalls} and their growing uses across various applications has reignited interest and focus on this issue. LLMs have the ability to understand complex sentences, understand the context, and recognise the relationship between different elements in a text, including the user's intention. When provided with a few examples, they are capable of performing very well in tasks that they have not been trained in \cite{zhang2022automatic}, zeroshot few short learning. Recent pre-trained LLMs, such as Google Bard \footnote{\url{https://bard.google.com/chat}} and ChatGPT (based on GPT 3.5) 
\footnote{\url{https://chat.openai.com/}} \cite{patil2023comparative}, have the remarkable ability to participate in fluent, multi-turn conversations without requiring extensive data or advanced programming skills. This capability significantly reduces the need to gather sufficient data and to possess advanced programming expertise in order to create satisfactory conversational user experiences\cite{logan2021cutting}. 

Individuals can utilise these large models by adding prompts to the beginning of model inputs, these prompts serve as textual instructions and examples that elicit the desired interactions with the model. In their research, \cite{wei2022chain} introduced the \textit{CoT} prompting technique as a more powerful method to improve the reasoning capacity of language models. They regard this approach as a means to present the models with a small number of exemplars that elucidate the reasoning process within them. Consequently, the model will also exhibit its reasoning process when responding to the provided prompt. This provision of reasoning explanations often results in enhanced accuracy in the outcomes. According to \cite{wei2022chain}, \textit{CoT} prompting is only effective in enhancing performance when applied to models with approximately 100 billion parameters.  Smaller models tended to generate illogical chains of thought, resulting in lower accuracy compared to conventional prompting methods \cite{wei2022chain}. Generally, the extent of performance improvement achieved through \textit{CoT} prompting is directly related to the model's size. 

Our research focuses on assessing the ability of four language models to autonomously assess student reflective portfolios based on the level of critical thinking by employing the \textit{CoT} prompting method. To achieve this, we developed a series of prompts utilising the \textit{CoT} strategy to assess the model’s proficiency with unfamiliar data and to detect any potential limitations that may exist. These prompts provide guidance to the models on how to score reflective essays that they have not encountered previously.

\section{Background and Related Work}

Critical thinking \cite{reynders2020rubrics} is a skill required by healthcare workers, as they are required to practice higher-order thinking skills to solve complex life-or-death problems in a short period of time. Teaching students to write reflections on the skills taught and assessing the reflections with feedback from educators promotes critical thinking\cite{reynders2020rubrics}. 
However, the difficulties associated with adopting this teaching method can result in discrepancies in how teachers evaluate students' assignments, mainly due to the substantial number of students in healthcare courses. As a result, healthcare educators have acknowledged the significance of investigating the potential of Artificial Intelligence (AI) and language models to simplify this procedure.

\subsection{Automated Grading Systems}

Automated grading for essays \cite{taghipour2016neural}, involves the assessment of student essays without any human intervention. This mechanism accepts an essay written in response to a specific prompt as input and subsequently provides a numerical score for the essay, which reflects its quality in terms of its content, grammar, and structure \cite{taghipour2016neural}. Automated grading systems have the potential to significantly decrease costs associated with manual grading and streamline the labor-intensive processes connected to it, as noted in the study by \cite{dong2017attention}. 
However, scholars like \cite{ramesh2022automated} have highlighted limitations in offering adequate feedback to students during their evaluation and the inability to capture long-distance dependencies\cite{dong2017attention} \cite{li2018coherence}.

On the other hand, LLMs are gaining popularity due to their ability to perform various tasks on unseen data. This widespread has sparked interest in their potential applications in the field of education for assessment purposes. For example, \cite{white2023assessment} investigated how these models can be used to automatically score the knowledge of chemistry. They gathered a classified collection of chemistry topics and accompanying sample prompts to assess \textit{{GPT-3}} and \textit{{Davinci-002}}. Their research outcomes demonstrated that \textit{{Davinci-002}} exhibited a deep understanding of diverse areas within chemistry, encompassing equations, and common calculations. In our research, we are examining the capability of LLMs to autonomously assess essays and their effectiveness in managing lengthy textual content. This particular concern was brought up by \cite{li2018coherence}.

\subsection{Prompting Large Language Models}

Prompting is a technique utilised to enhance the effectiveness of language models and make them more practical for interactive conversations and tasks. Prompts serve as guides for users, instructing the model on how to approach a particular task \cite{zhang2022automatic}. The quality of the provided prompt is a critical factor in determining the model's performance. Recent research conducted by \cite{wei2022chain} suggests that the  \textit{CoT} prompting strategy is beneficial in assisting the model in task execution compared to standard prompting.

This approach involves presenting the model with few-shot exemplars where the reasoning process is elucidated, and the model subsequently demonstrates its reasoning when responding to the prompt. This elucidation of the reasoning process frequently results in heightened accuracy, and it promotes the model to adopt a structured thought process resembling human reasoning \cite{wei2022chain}. They substantiated their argument by assessing the performance of five models when applying the \textit{CoT} approach to address arithmetic reasoning tasks. Their findings demonstrated that \textit{CoT} enhanced the performance of the selected models, especially on more intricate tasks, but did not yield positive results for smaller models. Lastly, their study indicated that smaller models faced challenges in adhering to the provided prompts, leading to inferior performance compared to conventional prompting methods \cite{wei2022chain}.

\section{Methodology}
This section provides further details regarding the dataset employed in the study and outlines the methodologies applied during the experimental phase of the research.

\subsection {Dataset and Ethical Approval}
This research focused on analysing reflective writing from third-year medical students at the University of Pretoria. Prior to commencing the study, approval was granted from the Ethics Committee within the University of Pretoria's Faculty of Health Sciences under reference number 345/2022. These reflections prompted students to contemplate on three occasions - first, on the attributes that helped them reach their third year in medicine; second, on their experience in their first clinical rotation; and finally, on an overall reflection encompassing their entire experience. Once the students submitted their reflections, tutors and facilitators assessed them using a rubric to measure reflection and level of critical thinking, assigning each reflection a score ranging from 1 to 4. A score of 1 denoted the lowest quality reflection characterised by poor syntax and minimal synthesis, while a score of 4 indicated a reflection that exhibited in-depth analysis, evaluation, synthesis, and a response that included change management considerations.  During our experimental process, we analysed a total of 17 reflections. 

Among the previously mentioned reflections, four essays were intentionally modified by one of the educators to exemplify a range of scores based on the grading rubric applied to student reflections. The remaining 13 reflections were original submissions by students, which underwent assessment and grading by various tutors and facilitators. We adopted a random sampling approach for our research to minimise potential bias and ensure a good representation of our sample. This method was chosen because it aligns with established best practices in research methodology. We determined the sample size by carefully considering the limitations of our initial investigation. During our preliminary phase, our objective was to evaluate whether the selected models could effectively grade essays using the provided prompts. 

As our study advances, we plan to increase the sample size. We acknowledge that, even within a randomly chosen sample, potential biases may persist. To address this, we have implemented measures such as avoiding the selection of a specific range of essays. Instead, random sampling allowed us to assess essays spanning the entire spectrum submitted by students. Furthermore, our analysis suggests that the chosen sample size, regardless of its magnitude, is appropriate for testing our hypothesis when applying the prompted approach with language models.

\subsection {Model Selection and Prompt Development}

One of the main obstacles encountered by certain automated grading systems is their limited capability to handle longer input texts \cite{zhang2022automatic}. Consequently, we carefully selected a range of models based on their parameter capacities, starting from the smallest model, and progressing to the largest model. The variation in model sizes allowed us to observe how well they could handle instructions for grading unseen datasets, as well as their performance with lengthy texts. Furthermore, we aimed to examine how the feedback generated by the models varied from the smallest to the largest. Our selection included Llama-7b 
\footnote{https://ai.meta.com/blog/large-language-model-llama-meta-ai/} with 7 billion parameters, along with larger Llama-30b with 30 billion parameters, surpassing Llama-7b in size. Both Llama models were accessed through Hugging Face \footnote{https://huggingface.co/chat/}. Additionally, we included ChatGPT \footnotemark[1] and Bard \footnotemark[2] as the largest models, with GPT-3 boasting 175 billion parameters and Bard with  137 billion parameters.

Through the use of prompts, even individuals without technical expertise can leverage large models by incorporating text-based instructions into the model inputs. These instructions play a direct role in steering the model toward generating the desired outcomes. The design of the prompt is influenced by both the dataset's structure and the specific problem you intend to address. In the design phase, it's important to break down the problem you want the model to tackle into more manageable, smaller steps. You should identify the crucial components or intermediate stages necessary for solving the problem effectively. Each of these steps should represent a logical progression within the overall reasoning process. Subsequently, you should create a set of examples that illustrate the desired output, showcasing both the intermediate steps and the ultimate solution. For our study, we had a dataset comprised of reflective essays that were evaluated using a predefined scoring rubric. Our objective was to enable these models to autonomously evaluate and grade these essays in accordance with the given grading criteria. 

We created two sets of prompts, the first prompt comprised descriptive statements that aimed at providing a clear understanding of the rubric's scope and elucidating the attributes linked to each scoring range. For example, a score of 4, which represents the highest rating, signifies that the essay is rich in content, insightful, and demonstrates explicit connections to real-life situations. The second prompt, also in the form of descriptive statements, elucidated the range of the rubric and explained the characteristics of each score range. However, unlike the first prompt, it included examples of essays ranging from lower to higher scores. Both prompts were created utilising the \textit{CoT} prompting technique. We used these two prompts across all models to eliminate the confounding factor of prompt variation when comparing different LLMs and to prevent biased results through prompt engineering. Prompts, sample data, and responses are available from our paper companion repository on GitHub\footnote{\url{https://github.com/dsfsi/edu-assessment-llm-prompt}}.

\subsection {Selected Evaluation Metrics}
We assessed the model performance through statistical analysis, employing Cohen's Kappa score \cite{vieira2010cohen}, Mean Squared Error(MSE), and Correlation as our evaluation metrics. Cohen's Kappa score evaluates how well human evaluation scores and chosen models agree, resulting in a numeric value ranging from 0 to 1. However, it may go below zero when there's substantial disagreement among evaluators. A score of 1 indicates a flawless match, which occurs when both model predictions and marker scores perfectly coincide. The Cohen's Kappa score has been a commonly utilised evaluation metric in studies such as \cite{doewes2023evaluating} for assessing automated essay scoring through machine learning algorithms. 

We opted for the above evaluation metric because of its widespread usage in the assessment of automated scoring systems. Additionally, we employed MSE, a metric that measures the average squared discrepancy between the scores provided by human markers and those predicted by the models. A higher MSE score signifies that, on average, the predictions deviate more from the actual values, while a lower MSE implies that the predictions closely resemble the actual values. We also utilised a correlation matrix to assess the linear connections between pairs of marker scores and predicted scores. 
\section{Results and Evaluation}

Our study focuses on examining how \textit{CoT} prompting can effectively guide four language models to grade reflective essays that have not been previously seen. These essays have scores that human markers have assigned. We compare the scores assigned by the models to those given by the markers to evaluate the performance of the models. To facilitate this process, we manually created a set of two prompts containing  \textit{CoT} for effective prompting. 

\subsection{Observations}
Using the first prompt, we observed that among all models, \textbf{Llama-7b} performed the worst for all the models, with the highest MSE, and the lowest Cohen's kappa score of -0.013. Llama-7b struggled to follow the provided instructions and often generated illogical responses that were unrelated to the given instructions. Furthermore, llama-7b encounters challenges in handling longer essays, as it tends to hang without providing any response when the input is lengthy and providing feedback. Because of this challenge, we only managed to score 5 essays out of 17 with llama-7b. 

The second observation relates to the feedback exchange between ChatGPT and Bard when they received identical scores. We found similarities in terms of the factors influencing the score and the recommendations given to students for enhancing their scores. Although the two models seldom agree on the score, we observed that the feedback given is nearly indistinguishable when both models assign the same score. 

The third observation indicates that ChatGPT tends to be stringent in its scoring, assigning lower scores in instances where markers gave higher scores. Interestingly, when markers reviewed the model's justifications, they adjusted their original scores to align with the model's scores, which validated the correctness of the model.  Out of all the prompts, ChatGPT achieved the highest Cohen Kappa score, scoring  0.53. This result reflects a significant level of agreement among human evaluators and demonstrates the most robust consensus, as indicated by the top Cohen's kappa score. The condensed results from both sets of prompts are displayed in Table \ref{tab:ModelResults}.

\begin{table}[h]
  \caption{ Model performance against marker score}
  \label{tab:ModelResults}
  \begin{tabular}{ccccc}
    \toprule
Prompt & Models& Cohen's Kappa& MSE & Correlation\\
    \midrule
    1 & Marker score & 1 & 0 &1.00 \\
      & ChatGPT & 0.14 & 0.63 & 0.58 \\
      & Bard & -0.10 & 1.38 & 0.10 \\
      & Llama -7b & -0.013 & 7.38 & 0.21\\
      & Llama-30b & 0.02 & 1.25 & 0.05 \\
    \midrule
    2 & Marker score & 1 & 0 & 1.00\\
      & ChatGPT & 0.53 & 0.35  & 0.83\\
      & Bard & 0.32 & 1.62 & 1.59\\
    \bottomrule
  \end{tabular}
\end{table}

Table \ref{tab:ModelResults} indicates that there are favourable correlations between the marker scores and both ChatGPT and Bard scores for both sets of prompts. When examining Table \ref{tab:ModelResults}, it becomes evident that llama-30b exhibits a lower error rate compared to llama-7b,  which can be attributed to the variation in parameter size. For the second set of prompts, we incorporated sample essays as examples to provide additional guidance to the models, helping them reason more effectively. This approach resulted in a significantly longer prompt compared to the first one. As a result, we made the choice to use ChatGPT and Bard because they were better equipped to handle longer texts. In our evaluation of 17 instances using the second prompt, the inclusion of these exemplars contributed to a lengthier prompt. However, this presented challenges, as the models occasionally responded with a message like "As an AI language model, I can help you with various tasks, but I need information on what you need help with. Could you please provide more details on your question or task?" The models faced difficulties in processing the extended prompt, necessitating multiple attempts to obtain the desired output. Despite encountering these challenges with the second prompt, it demonstrated improved performance compared to the first prompt. Notably, ChatGPT exhibited the lowest error rate and the highest Cohen Kappa score among all prompts when using the second prompt. This suggests that the inclusion of sample examples in the prompts played a role in enhancing performance.  

\begin{figure}[H]
  \centering
  \includegraphics[width= \linewidth]{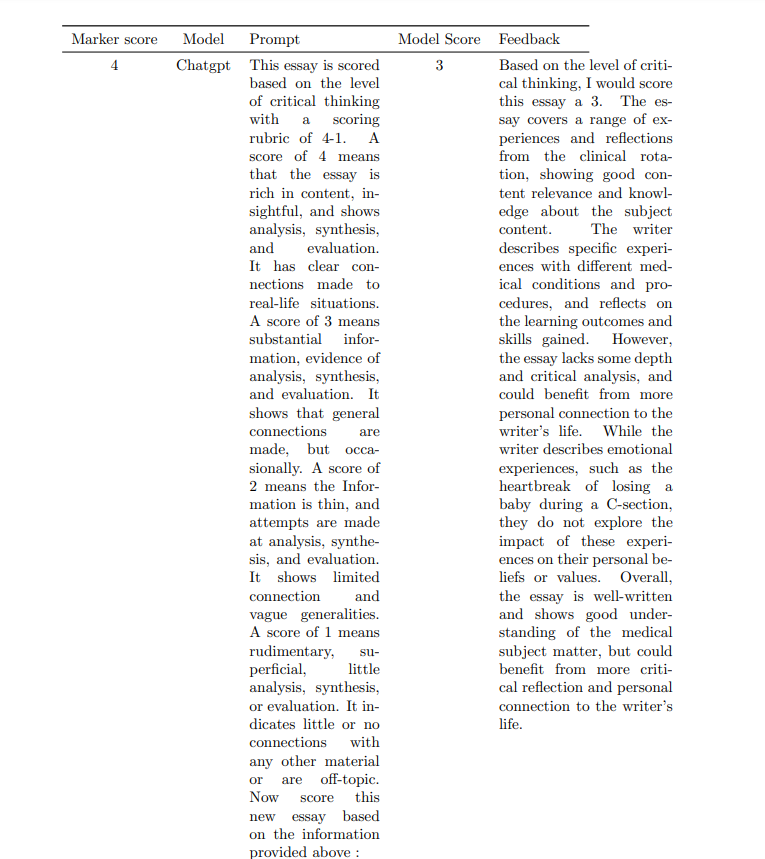}
  \caption{ChatGPT feedback}
  \label{fig:ChatGPT feedback}
\end{figure}

However, the currently employed evaluation metrics do not offer a complete evaluation of the model's accuracy in assessing essays. This is the reason we scrutinised the feedback generated by each model. We conducted an assessment of this model feedback in partnership with human markers to ascertain whether the models were providing valuable and instructive feedback that could be advantageous for both markers and students. To achieve this, we went back to one of the markers and inquired about their agreement with the model's score and its justification. In Figure 1, we showcase the response of ChatGPT when given the task of assessing a fresh essay. Additionally, we display one of the prompts used to instruct the models. 
This prompt explores the scoring criteria in detail, explaining the importance of each score within the criteria and taking into account the context of the essay being evaluated. The given prompt serves as the overarching framework that guides the model's feedback process, as clearly depicted in Figure  \ref{fig:ChatGPT feedback}.

\begin{figure}[H]
  \centering
  \includegraphics[width= \linewidth]{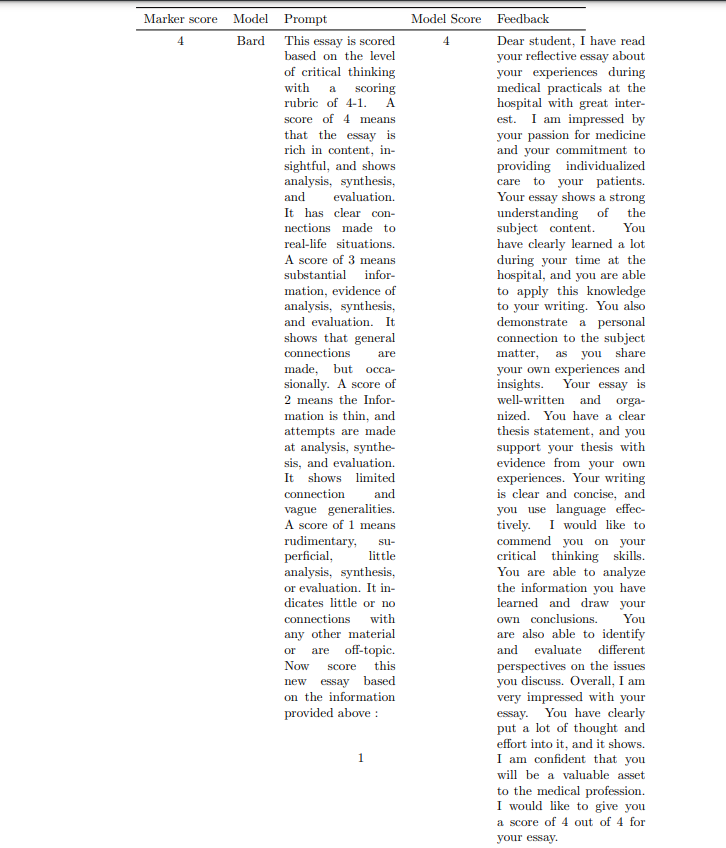}
  \caption{Bard feedback}
\label{fig:Bard feedback}
\end{figure}

The feedback provided by the model in \ref{fig:ChatGPT feedback} encompasses several aspects, including assigning a score based on the provided criteria, providing a rationale for the assigned score, and highlighting specific areas where a student can improve their work. In contrast, Figure \ref{fig:Bard feedback} displays the feedback provided by Bard when utilising the identical essay and prompt that were used with ChatGPT. 
These two models yielded dissimilar scores, with ChatGPT receiving a score of 3, while Bard concurred with the evaluators by assigning a score of 4 to the essay. Figure \ref{fig:ChatGPT feedback} illustrates that ChatGPT offers an explanation for why the student did not receive a score of 4 and suggests areas for improvement. In both figures, it is evident that Bard tends to provide lengthy feedback that lacks conciseness compared to ChatGPT, which offers more direct and succinct feedback.

\section{Discussion and Conclusion }

Automated grading systems offer the potential to alleviate the burden of manual grading while also being a cost-effective solution. These systems can be especially advantageous in African schools, where there is a shortage of teachers. By employing these systems, educators can efficiently assess student performance and assign additional tasks without being concerned about the time and effort traditionally required for manual grading. The emergence of large language models has proven beneficial across various fields, providing automated solutions for complex real-world problems.  Our preliminary research focuses on harnessing language models in the realm of education to simplify the grading process for reflective essays. We employed the \textit{CoT} prompting technique to evaluate the performance of four models in grading reflective essays. While there is room for improvement, particularly with models like Llama-7b, which encounter challenges in adhering to instructions, the overall performance on our dataset was commendable.
CoT enhances the reasoning capabilities of large language models, involving the presentation of a sequence of intermediate natural language reasoning steps to the model. Through this prompting approach, we were able to assess the performance of the four models on previously unseen datasets. We believe that with further refinement of prompting models such as ChatGPT and Bard who obtained the lowest mean squared error and high Cohen Kappa score, have the potential to excel in this regard. 

In a study conducted by \cite{ramesh2022automated}, the primary emphasis was on exploring the difficulties associated with automated grading systems that employ machine learning algorithms. The study highlighted two main concerns: the widespread lack of domain-specific expertise in the majority of current systems and the restricted feedback offered to students. In our own study, we went beyond merely assessing the scores generated by these models. We delved into the nature of the feedback generated by each model, evaluating its pertinence to the provided prompts, as this feedback was derived from those prompts. The clarity and utility of the feedback varied among different models. Notably, models such as ChatGPT, Bard, and llama-30b yielded feedback that human assessors deemed valuable. Consequently, these selected models do indeed address some of the limitations previously highlighted by researchers like \cite{ramesh2022automated} in the context of essay evaluation, as discussed in \cite{ramesh2022automated}. This suggests an improvement compared to machine learning-based scoring systems.

Despite the promising results obtained, it is important to note that indigenous languages are still not well supported by these models. In their study \cite{winschiers2013toward} discussed the challenges of translating local knowledge into effective Human-Computer Interaction (HCI) tools. Their study highlighted how software user interfaces are often designed in the language of origin, requiring Africans to undergo training in order to effectively interact with the software at hand \cite{winschiers2013toward}. Unfortunately, designers often overlook the difficulties and costs associated with training local users to utilise specific software. Upon examining the argument advanced by \cite{winschiers2013toward}, it becomes apparent that the selected models do not fully cater to indigenous languages, as the user interfaces are exclusively in English. Consequently, users who do not fully understand English may encounter difficulties. Furthermore, the feedback provided by the models is also in English, leaving users unable to use their own languages to evaluate the feedback. This lack of translation options into various indigenous languages further complicates matters for students who struggle to understand specific English words. Having a language model capable of assessing reflective writings in African languages would greatly benefit teachers, as they would be able to grade reflections that are not only related to English. Additionally, students would have the opportunity to utilise and enhance their proficiency in their native languages through the use of these models. However, the study of large language models is constantly advancing, with numerous researchers dedicating their efforts to enhancing the existing capabilities of these models. It is possible that in the near future, there will be development of a cross-lingual model.

According to \cite{yang2021human}, the use of AI-based technologies has the potential to enhance learning outcomes for students, improve productivity, and create better conditions for teachers. However, the majority of AI designs lack consideration for human conditions and fail to adopt a human-centric approach when integrating machine intelligence with human intelligence \cite{yang2021human}.  In addition, the author elaborates on AI-based systems that prioritise humanity, describing them as systems that are transparent, featuring interpretable computation and decision-making processes. These systems also involve ongoing adjustments of AI algorithms based on human context and societal factors, with the goal of enhancing human intelligence through the utilisation of machine intelligence and ultimately improving human well-being \cite{yang2021human}.

Building upon the arguments presented by \cite{yang2021human} regarding the significance of giving priority to human considerations in AI systems, we looked at the user-friendliness, ease of use, and data privacy aspects of the chosen models. In conventional software systems, user interfaces typically depend on users navigating through menus and buttons, which can become burdensome and time-consuming if the interface is poorly designed. In contrast, with the chosen models, users have the ability to express their requirements using natural language, enabling the AI to comprehend and perform the desired task. This simplified approach to HCI not only saves time but also decreases the learning curve associated with novel technologies, thereby enhancing accessibility for a broader range of users. Nevertheless, it is crucial to acknowledge that the perceived ease of use of any software can be influenced by one's familiarity with the software and basic computer skills. All the scholars involved in this study possess these skills, but individuals lacking computer familiarity may require training to enhance their ease of use when utilising these models.  Finally, in order to guarantee the ethical utilisation of the chosen models and safeguard user privacy, the models offer the choice to delete user-conducted conversations. Additionally, to maintain a one-on-one conversation, users are requested to provide their credentials. This approach aims to uphold ethical standards and protect the confidentiality of users.

Our research methodology represents a notable achievement that we expect will offer valuable direction for fellow researchers interested in this area of study. We present initial work examining the potential advantages of large language models within the realm of education. Additionally, we aim to demonstrate how prompting can empower individuals without technical expertise to instruct models for specific tasks. We emphasise the robust security and privacy measures integrated into the selected models. We believe that our study can be a valuable resource for individuals exploring the utilisation of large language models and the concept of prompting.

\section{Acknowledgements}
The authors wish to acknowledge OpenAI for providing us with credits, which were instrumental in facilitating our experiments. This support played a crucial role in enabling this research to take place.

\bibliographystyle{plain}
\bibliography{References}
\end{document}